\documentclass[letterpaper, 10 pt, conference]{ieeeconf}  % Comment this line out if you need a4paper
\pdfoutput=1

\IEEEoverridecommandlockouts                              % This command is only needed if 
\usepackage{graphics} % for pdf, bitmapped graphics files
\usepackage{epsfig} % for postscript graphics files
\usepackage{times} % assumes new font selection scheme installed
\usepackage{amsmath} % assumes amsmath package installed
\usepackage{amssymb}  % assumes amsmath package installed

\usepackage{cite}
\usepackage{multirow}
\usepackage{makecell}

\usepackage{etoolbox}
\makeatletter
\patchcmd{\@makecaption}
  {\scshape}
  {}
  {}
  {}
\makeatletter
\patchcmd{\@makecaption}
  {\\}
  {.\ }
  {}
  {}
\makeatother

\usepackage{float}

\title{\LARGE \bf
Learning Excavation of Rigid Objects with Offline Reinforcement Learning
}

\author{Shiyu Jin$^{1}$, Zhixian Ye$^{1}$ and Liangjun Zhang$^{1}$% <-this % stops a space
\thanks{$^{1}$The authors are with Robotics and Autonomous Driving Lab, Baidu Research, USA. {\tt\small \{shiyujin, zhixianye, liangjunzhang\}@baidu.com}}%
}

\begin{document}

\maketitle
\thispagestyle{empty}
\pagestyle{empty}

%%%%%%%%%%%%%%%%%%%%%%%%%%%%%%%%%%%%%%%%%%%%%%%%%%%%%%%%%%%%%%%%%%%%%%%%%%%%%%%%
\begin{abstract}
Autonomous excavation is a challenging task. The unknown contact dynamics between the excavator bucket and the terrain could easily result in large contact forces and jamming problems during excavation. Traditional model-based methods struggle to handle such problems due to complex dynamic modeling. In this paper, we formulate the excavation skills with three novel manipulation primitives. 
We propose to learn the manipulation primitives with offline reinforcement learning (RL) to avoid large amounts of online robot interactions. The proposed method can learn efficient penetration skills from sub-optimal demonstrations, which contain sub-trajectories that can be ``stitched" together to formulate an optimal trajectory without causing jamming. We evaluate the proposed method with extensive experiments on excavating a variety of rigid objects and demonstrate that the learned policy outperforms the demonstrations. We also show that the learned policy can quickly adapt to unseen and challenging fragmented rocks with online fine-tuning.
\end{abstract}

%%%%%%%%%%%%%%%%%%%%%%%%%%%%%%%%%%%%%%%%%%%%%%%%%%%%%%%%%%%%%%%%%%%%%%%%%%%%%%%%
\section{INTRODUCTION}
\begin{figure}
	\centering
    \includegraphics[scale=0.38]{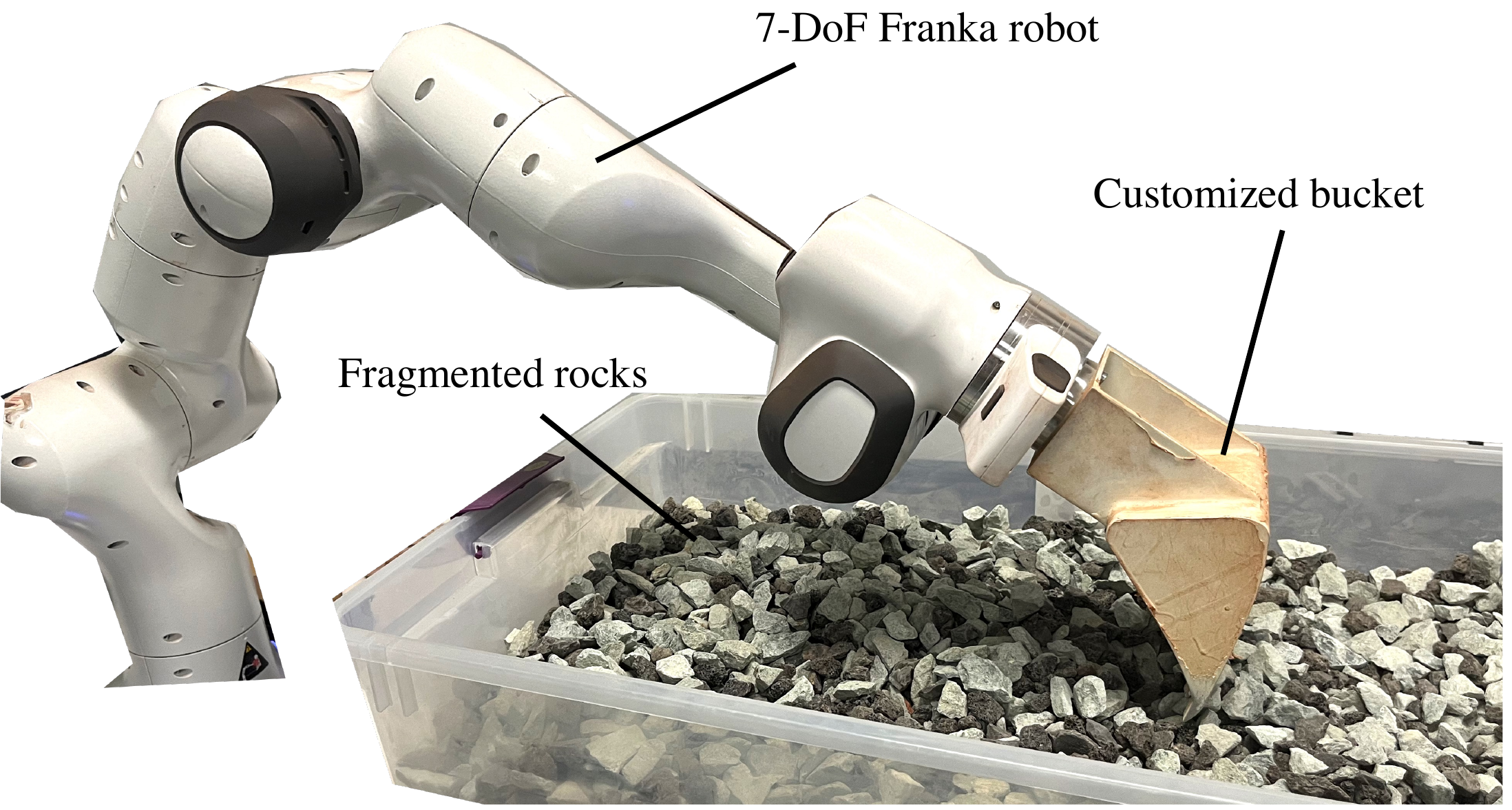}
    \caption{Excavation of rigid objects. A robotic manipulator attempts to penetrate the end-effector bucket into the fragmented rocks without causing jamming.} 
    \label{fig:setup}
\end{figure}

Excavators have been widely used in mining and construction environments for many decades. The excavator operations usually rely on experienced operators so that they can choose appropriate actions for different tasks. While we have seen a lot of studies in the field of robotics and autonomous driving, excavators can achieve only limited autonomy \cite{zhang2021autonomous, jelavic2021towards, jud2019autonomous}. One of the main difficulties lies in the fact that excavation tasks usually require rich contact between the excavator bucket and the terrain. The problem becomes even harder if the terrain being excavated contains irregular rigid objects, such as fragmented rocks (Fig. \ref{fig:setup}), where the contact dynamics are complex and extremely hard to model. In such scenarios, improperly applied forces could easily result in jamming problems between the bucket and the terrain \cite{lu2021excavation_rigid, lu2022excavation, zhu2022excavation}. Handling rigid objects is crucial for excavation in mining and construction environments, where the excavators often have to deal with large rocks and stones. But there has been relatively little work examining how excavators can achieve such objectives autonomously.

Many previous works have studied the planning \cite{zhao2021tasknet, %yang2019compact, 
yang2021optimization, 
%zhixianIROS, 
jelavic2019whole, jelavic2020terrain} and the control \cite{fernando2021control, jud2017planning, lee2022precision, park2016online, sotiropoulos2021dynamic, sotiropoulos2019model, sandzimier2020data} for the excavation of granular materials, such as sands. The granular materials are relatively easy to be excavated because of the large torques that can be generated from the hydraulic system in excavators. In contrast, irregular rigid objects, such as fragmented rocks and stones, which cannot be disintegrated during excavation, are extremely challenging to be excavated. Only a few works have studied the autonomous excavation of irregular rigid objects. Lu et al. \cite{lu2021excavation_rigid, lu2022excavation} learn the excavation of wood blocks in simulation and then transfer it to the real world. Zhu et al. \cite{zhu2022excavation} learn the dynamic model for wood blocks excavation. However, due to the huge sim-to-real gap and the bias in the learned dynamic model, those works can only achieve low excavation success rates on a single type of terrain, wood blocks. A lot of jamming problems between the bucket and the rigid objects cannot be properly handled by the previous works. 

To the best of our knowledge, there is no previous work that can efficiently avoid jamming during the excavation of rigid objects. And there are also no methods that can generalize the excavation skills across different rigid object types. We are the first to conduct a study on the autonomous excavation of rigid objects covering a broad range of types, including red mulch, marble chips, and fragmented rocks, etc. 
Although Egli et al. \cite{egli2022soil} address different soil excavations through domain randomization using a soil analytical model, the soil analytical model is only applicable to homogeneous soils instead of irregular rigid objects, where the contact dynamics are much more complex. 

In this paper, we propose to learn rigid object excavation with offline reinforcement learning. Reinforcement learning algorithms have shown the ability to learn an efficient policy for robotic manipulation tasks with good generalization abilities. But a large number of on-robot trails make it difficult to be deployed in real-world contact-rich robotic manipulation tasks, such as excavations. In contrast, offline RL can avoid online robot interaction by learning the policy with previously collected datasets. For the excavation of rigid objects, the offline dataset is obtained from teleoperation and scripted policy, which avoids dangerous actions during on-robot RL training. To efficiently penetrate the terrain of the rigid object while avoiding the jamming problems, we design three manipulation primitives, \texttt{sweep}, \texttt{rotate}, and \texttt{penetrate} based on expert experiences. The \texttt{sweep} and \texttt{rotate} primitives generate horizontal bucket motion and rotation in the pitch axis to remove the small fragments that block the bucket penetration. The \texttt{penetrate} primitive generates vertical bucket motion to penetrate into the terrain. 
%Each primitive consists of continuous parameters for the reference bucket velocity as well as the force and displacement limits. 
The parameters of the manipulation primitives are learned as the policy output using offline RL.
To generalize across different terrain configurations and terrain types, we propose to utilize two long short-term memory (LSTM) encoders. 
% One encoder extracts the excavation progress and the current terrain configuration from the current excavation trajectories. The other encoder learns the terrain types and the information necessary to complete the task from the previously collected excavation trajectories. 
Experiments show that the proposed method is able to learn a better excavation policy than the demonstrations, which contain sub-optimal excavation trajectories. Our proposed framework also has the ability to quickly adapt to unseen rigid objects with a few online fine-tuning.

The main contributions presented in this work are summarized below:
\begin{itemize}
% \item We are the first to study the challenging jamming problems for the excavation of irregular rigid objects.
\item To the best of our knowledge, we are the first to study the jamming problems for the excavation of a variety of rigid objects, which would be beneficial to the robotics and autonomous construction community.
\item We proposed a framework for the excavation of rigid objects with offline reinforcement learning. The proposed method can learn excavation policy from sub-optimal demonstration trajectories.
\item We designed multiple manipulation primitives with continuous parameters which, once trained, are able to generate robot commands to excavate rigid objects without causing jamming problems.
\item We empirically evaluated the performance of our method by extensive experiments. The learned policy outperforms the sub-optimal demonstrations. In addition, the learned policy is able to quickly adapt to unseen and challenging fragmented rocks with online fine-tuning.
\end{itemize}

% We have several modifications. Instead of manually selecting a fixed length of past trajectory, we encode the past trajectories in latent variables using LSTM, which captures the knowledge of the terrain states at the current time step. Instead of selecting a discrete primitive from a primitives pool, we learn continuous parameters for each primitive and the action can be a combination of different primitives with continuous parameters. We also add a context for different tasks which allows the generalization across tasks. By utilizing the Offline RL, our method can learn better penetration policy even if the demonstrations in offline dataset are sub-optimal.

\section{RELATED WORK}
\subsection{Planning and Control for Autonomous Excavation}

In recent years, a lot of works have been done in Autonomous Excavation. Some works focus on building an integrated system. For example, Zhang et al. \cite{zhang2021autonomous} propose an autonomous excavator system for material loading tasks, which can continuously operate for 24 hours without any human intervention; Jelavic et al. \cite{jelavic2021towards} present an integrated system for performing precision harvesting missions, which combines mapping, localization, planning, and control; Jud et al. \cite{jud2019autonomous} achieve autonomous free-form trenching using a walking excavator. Many other works focus on the task or manipulator trajectory planning for the excavator using optimization-based or data-driven methods \cite{zhao2021tasknet, 
%yang2019compact, zhixianIROS, 
yang2021optimization, jelavic2019whole, jelavic2020terrain}. Since excavation is a contact-rich robotic manipulation task, simply executing a planned trajectory could easily fail due to the unknown contact dynamics between the terrain and the excavator bucket. So some works focus on autonomous excavator control \cite{fernando2021control, jud2017planning, lee2022precision, park2016online, sotiropoulos2021dynamic, sotiropoulos2019model, sandzimier2020data}. The target for most of the excavation tasks is usually to collect the granular objects. Only a few works study the excavation of irregular rigid objects, where the contact dynamics are much more complex than the excavation of granular objects. Sotiropoulos et al. \cite{sotiropoulos2020autonomous, sotiropoulos2021methods} utilize Gaussian Process and Unscented Kalman Filter to capture a rock. The authors have to facilitate the rock collection by putting sand underneath the rock. Lu and Zhang \cite{lu2021excavation_rigid} train a classifier for wood blocks excavation using voxel-based representation. But they achieve low excavation success rates due to the jamming problems.

Reinforcement learning has demonstrated its ability to solve some contact-rich robotic manipulation tasks \cite{gu2017deep, rajeswaran2017learning,zhang2022learning }. Previous works also show the RL applications in autonomous excavation. Hodel et al. \cite{hodel2018learning} and Kurinov et al. \cite{kurinov2020automated} learn to load an excavator in simulations using RL. But they do not perform any experiments in the real world. Egli et al. \cite{egli2020towards, egli2022general} learn a tracking controller by collecting data from the real-world operation of a hydraulic excavator arm. But the learned RL policy only works for free space excavator operation without any interaction with the terrain. They \cite{egli2022soil} also learn a controller for a full-sized hydraulic excavator that can adapt online to different soil characteristics. The RL policy is trained in simulation and deployed directly in the real world. But the method requires an analytical soil model, which cannot be generalized well if the terrain type changes to rigid objects where an analytical contact model is hard to obtain. Lu et al. \cite{lu2022excavation} and Zhu et al. \cite{zhu2022excavation} utilize RL to learn the excavation of wood blocks. Although they achieve successful excavation of rigid objects with geometric representation \cite{lu2022excavation} and learned contact dynamic model \cite{zhu2022excavation}, their methods still suffer a lot of failure cases where jamming happens during bucket penetration. The MPC controller in their method is also time-consuming to compute. 

\subsection{Offline Reinforcement Learning in Robotic Manipulation}

Different from the above, we directly learn the excavation policy from the offline dataset, which does not require dangerous on-robot RL training. Several prior works have explored offline RL methods for learning robotic grasping and manipulation skills \cite{levine2020offline, pinto2016supersizing, dasari2019robonet, mandlekar2020learning, mandlekar2020iris, zhao2022offline}. Nair et al. \cite{nair2020awac} learn a policy that maximizes reward while bounding the deviation
from the dataset. But the proposed method tends to overfit heavily with many offline gradient steps. Kostrikov et al. \cite{kostrikov2021offline} propose offline reinforcement learning with Implicit Q-Learning (IQL) to estimate the maximum Q-value over actions that are in support of the data distribution using the Expectile Regression. Experiments have shown that IQL performs well on tasks that require multi-step dynamic programming. This is a good fit for the excavation tasks, where we are trying to find an optimal penetration policy by ``stitching" several sub-optimal trajectories from the demonstrations. In this work, we build on IQL to learn a policy for the excavation of rigid objects.

% In order to deal with the jamming problem during bucket penetration, we propose to learn bucket penetration policy from demonstration. By introducing manipulation primitives, bucket can avoid jamming during penetration. By learning from offline dataset, penetration policy can be learned efficiently without online interaction with the environment. By using offline RL, we can learn a penetration policy that outperforms the sub-optimal demonstration. Some state of the art offline RL, \cite{nair2020awac, kostrikov2021offline, zhao2022offline}. 

\section{Background and Problem Formulation}
\subsection{Offline Reinforcement Learning}
The reinforcement learning problem is formulated using a Markov Decision Process (MDP) associated with task $\mathcal{T}$ defined by $\mathcal{M}_\mathcal{T} =
\{\mathcal{S}, \mathcal{A},\mathcal{P}, r, p_0(s), \gamma \}$, where $\mathcal{S}$ is the state space, $\mathcal{A}$ is the action
space, $\mathcal{P}$ is the environmental dynamics, $r$ is the reward function, $p_0(s)$ is the initial state distribution, and $\gamma$ is the discount factor. The agent interacts with the MDP according to a policy $\pi (a|s)$. The final goal in an RL problem is to learn a policy that maximizes the cumulative discounted rewards $ R = \sum_{t=0}^{T} \gamma ^t r(s_t, a_t)$.

Offline reinforcement learning problem can be defined as a data-driven formulation of the reinforcement learning problem. The end goal is still to optimize the RL objective. However, the agent no longer has the ability to interact with the environment and collect additional transitions using the behavior policy. Instead, the learning algorithm is provided with a static dataset of transitions, $\mathcal{D} = \{(s_t^i, a_t^i, s_{t+1}^i, r_t^i)\}$, and must learn the best policy it can using this dataset \cite{levine2020offline}. 

\subsection{Problem Formulation}
\label{problem_formulation_section}
\begin{figure}[H]
	\centering
    \includegraphics[scale=0.43]{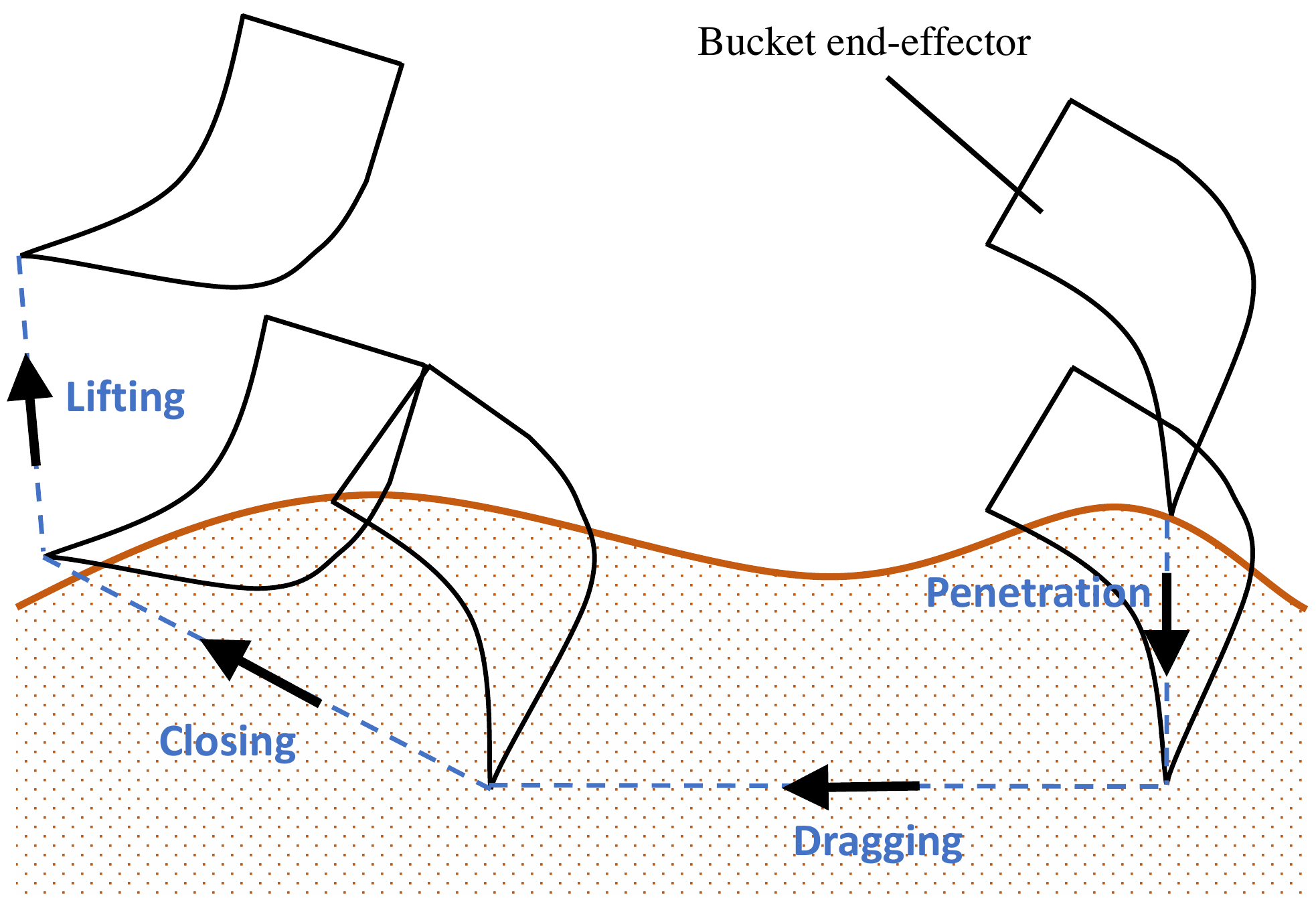}
    \caption{4 phases in the excavation process: penetration, dragging, closing, and lifting. The black arrows represent the bucket movement in each phase. In this work, we only focus on the penetration phase, where the jamming problem often happens. The brown line shows the terrain surface.} 
    \label{fig:phases}
\end{figure}

\begin{figure}[H]
	\centering
    \includegraphics[scale=0.53]{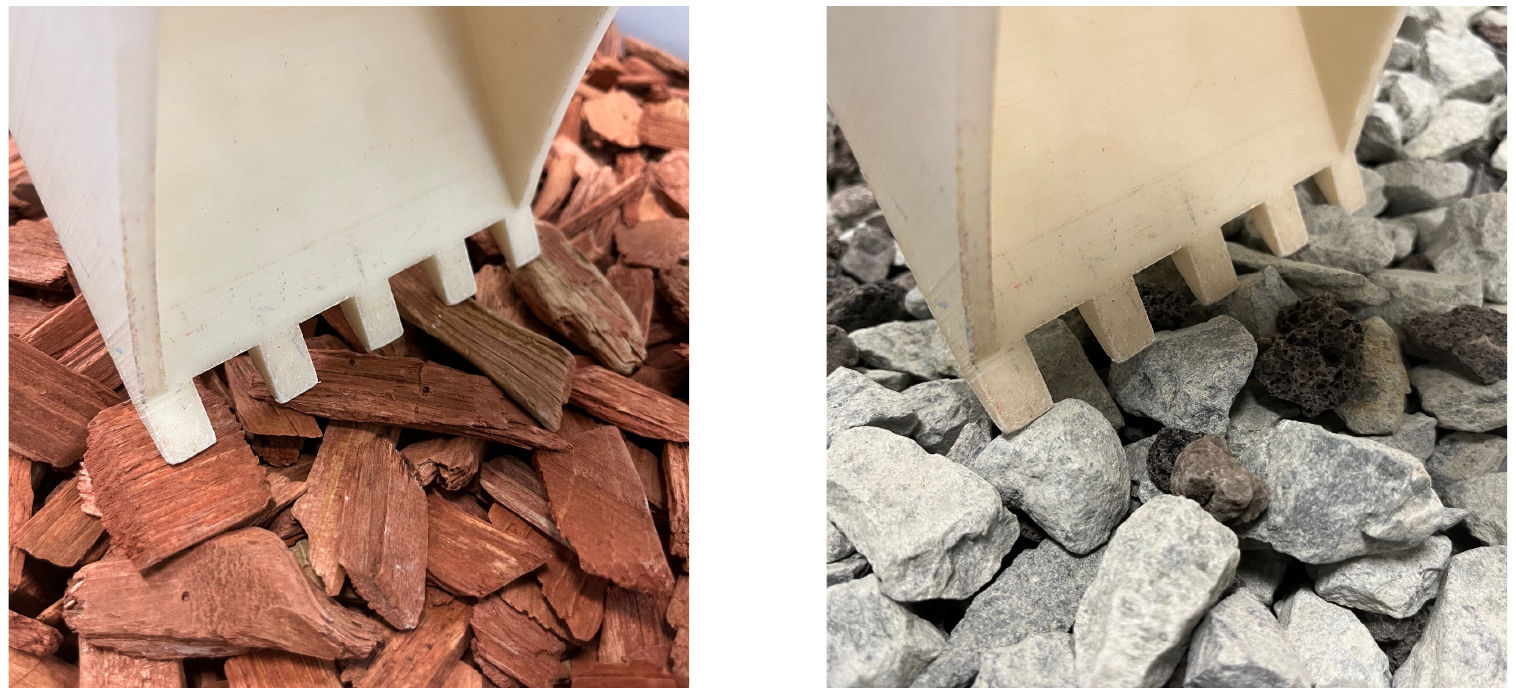}
    \caption{Vertically downward bucket movements will result in large contact forces and jamming. The bucket cannot penetrate into the terrain. An optimal strategy should sweep the surface of the terrain for downstream penetration.} 
    \label{fig:jamming}
\end{figure}

In the autonomous excavation literature, the excavation process can usually be divided into 4 phases: penetration, dragging, closing, and lifting (Fig. \ref{fig:phases}) \cite{zhu2022excavation}. The jamming problem often happens at the penetration phase during the excavation of irregular rigid objects. Compared to the granular material, such as sand, irregular rigid objects are larger in size and non-deformable. The contact force between the bucket and the terrain can easily become large and result in jamming (Fig. \ref{fig:jamming}), where the bucket cannot follow the planned path due to the contact with the terrain. In this work, we focus on how to penetrate the excavator bucket efficiently without causing jamming. The study of the other three phases is left to future work.

We formulate a robotic excavation penetration of rigid objects problem. Consider a flat terrain composed of irregular rigid objects, a robotic manipulator attempts to penetrate the end-effector bucket into the terrain at a certain depth for downstream excavation. Based on the bucket pose, the bucket velocity, and the measured contact force, the agent needs to find the appropriate robot command to penetrate into the terrain. We make the following assumptions on the task: 1) we emulate an excavator using a $7 DoF$ Franka manipulator with a bucket as the end-effector; 2) the objects being excavated are reachable by the bucket; 3) if the teeth of the bucket reach a certain depth $d_{target}$ below the surface of the terrain, we assume that the penetration phase is finished; 4) when a large contact force is detected, the robot controller would take over and issue a halt to the robot, which we use as a criterion for whether jamming happens or not throughout the paper; 5) after the penetration phase, the rest of the trajectory including dragging, closing, and lifting can be efficiently planned so that no jamming will happen in those phases. 

% Based on our experience, if the excavation trajectory is well planned, the jamming problem will not happen at the dragging, closing, and lifting phase. But simply following a fixed trajectory could easily resulting in jamming in the penetration phase. \cite{yifanIROS} tries to learn the dynamic model. But their method still have about $25\%$ jamming even for shallow excavation of wood building blocks. And their method will not work for terrains like red mulch and small rocks. In this work, we modify the primitives design to make it work for different kinds of materials. With our primitives, the penetration success rate can reach $100\%$ without jamming. We also learn the penetration policy using offline RL, so that the policy can be generalized to different terrains. We make the following assumptions on the task: 1) we emulate a $4 DoF$ excavation with a $7 DoF$ Franka manipulator with a bucket as the end-effector. The terrain being excavated is reachable by the bucket. 2) if the teeth of the bucket is penetrated $d$ below the surface of the terrain, we assume that the penetration phase is finished. We empirically find that $d=5 cm$ is deep enough for excavation. 3) after the penetration phase, the rest of the trajectory including dragging, closing, and lifting can be well planned so that no jamming will happen in those phases. 

\section{Approach}
We propose a robotic excavation penetration framework for efficiently penetrating different terrains. We design three manipulation primitives to avoid jamming during the penetration. The parameters of the manipulation primitives are learned using offline RL from a pre-collected dataset. The dataset consists of demonstrated penetration trajectories collected using teleoperation and scripted policy. With IQL, we can learn the penetration policy even if the penetration trajectories in the offline dataset are sub-optimal. Intuitively, IQL stitches several sub-optimal penetration trajectories and finds an optimal penetration policy that is not only without jamming, but also fast and with smaller contact forces. To generalize the learned penetration policy, we introduce two LSTM auto-encoders to encode the terrain configurations and terrain types using the current excavation trajectories and the demonstrated excavation trajectories, respectively. Given a new terrain type, the penetration policy can be fine-tuned with a few online data collections.

% The goal is to penetrate the bucket a certain depth $d$ below the surface of the terrain. We define several manipulation primitives as the bucket desired motions. We then learn the penetration policy using offline reinforcement learning from a demonstrated penetration dataset. The demonstration dataset contains bucket penetration trajectories from random action, teleoperation, and scripted policy. With IQL \cite{kostrikov2021offline}, we can learn optimal penetration policy even if the demonstration dataset only contains sub-optimal penetration trajectories. Intuitively, we would like to learn a bucket penetration RL policy that can penetrate the terrain not only without jamming, but also fast and with small contact force.

\subsection{Manipulation Primitives for Excavation Penetration}
We define the manipulation primitives as the bucket's desired movement. The desired movement is a reference bucket velocity tracked by an impedance controller. In \cite{zhu2022excavation}, only discrete actions are empirically selected as the primitives. The limited choices of 10 discrete primitives could result in sub-optimal and less flexible policies. And the primitives that work for one rigid object may not be able to generalize to other rigid objects.

\begin{figure}
	\centering
    \includegraphics[scale=0.41]{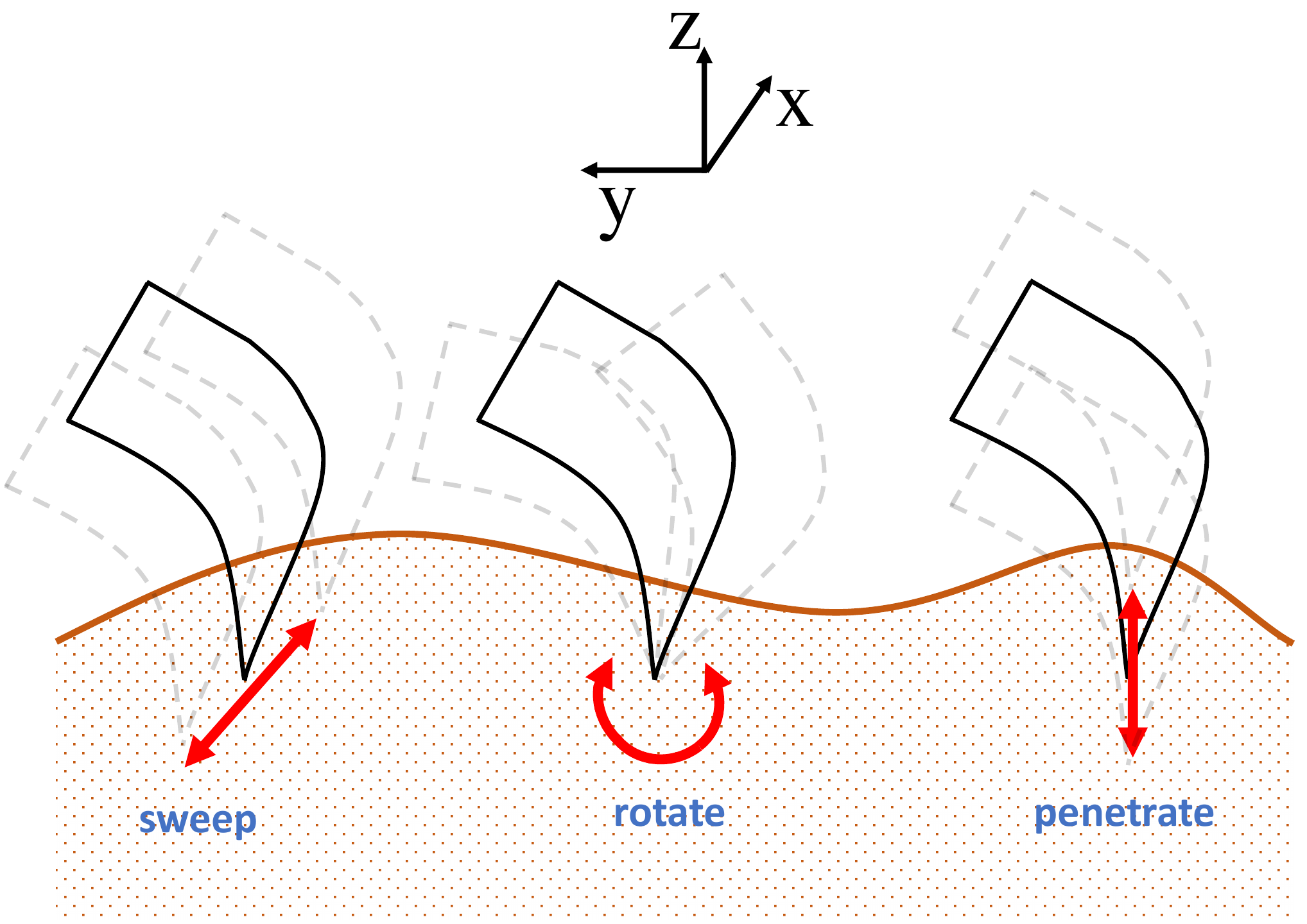}
    \caption{3 manipulation primitives: \texttt{sweep}, \texttt{rotate}, and \texttt{penetration}. In \texttt{sweep}, the robot generates horizontal bucket motion in the $x$ axis; in \texttt{rotate}, the robot generates rotational bucket motion in the pitch axis; in \texttt{penetrate}, the robot generates vertical bucket motion in $z$ axis. \texttt{sweep} and \texttt{rotate} manipulate the surface of the terrain to facilitate the penetration, and \texttt{penetrate} selects proper vertical velocity to penetrate the terrain. The red arrows represent the desired bucket movement in each primitive.} 
    \label{fig:primitives}
\end{figure}

Different from the discrete actions in \cite{zhu2022excavation}, we propose to use three primitives with continuous parameters, \texttt{sweep}, \texttt{rotate}, and \texttt{penetration} (Fig. \ref{fig:primitives}). With the continuous parameters, we are able to select continuous bucket velocities and force/displacement limits in order to avoid jamming. Each primitive generates a bucket velocity in one dimension with the reference frame shown in Fig. \ref{fig:primitives}. Specifically, in \texttt{sweep}, the robot generates horizontal bucket motion in the $x$ axis, so that the teeth of the bucket will sweep the surface of the terrain and remove the fragments that cause jamming. In \texttt{rotate}, the robot generates rotational bucket motion in the pitch axis to loose the surface of the terrain. In \texttt{penetrate}, the robot generates vertical bucket motion in the $z$ axis to penetrate the bucket into the terrain. The parameters of \texttt{sweep} and \texttt{rotate} are $(v^x, F_{lim}^x, d_{lim}^x)$ and $(\omega^{pitch}, M_{lim}^{pitch}, \alpha_{lim}^{pitch})$, respectively, where $v^x$ and $\omega^{pitch}$ are continuous reference velocities, $F_{lim}^x$ and $M_{lim}^{pitch}$ are force/torque limits, and $d_{lim}^x$ and $\alpha_{lim}^{pitch}$ are displacement/angle limits. The limits are the threshold where the primitives should change the moving direction. The parameters of \texttt{penetrate} are $(v^z, F_{lim}^z)$. If the force limit $F_{lim}^z$ is reached, the bucket will be lifted up to avoid jamming. 

Intuitively, \texttt{sweep} and \texttt{rotate} manipulate the surface of the terrain to facilitate the penetration, and \texttt{penetrate} selects proper vertical velocity to penetrate the terrain. A small force may not generate enough force to manipulate the terrain surface or penetrate the terrain, while a large force could easily result in jamming.

% In order to avoid jamming, we introduce a \texttt{sweep} and a \texttt{penetrate} primitive. The \texttt{sweep} generates horizontal bucket motion, so that the teeth of the bucket will sweep the surface of the terrain and remove the small fragments which causes jamming. The \texttt{penetrate} primitive generates vertical bucket motion, so that the bucket can penetrate the terrain and reach a certain depth below the surface. Each primitive consists of a continuous parameter representing the reference bucket velocity for the impedance controller to track. For the \texttt{sweep} primitive, a large parameter could result in large contact force in the sweep direction, which causes jamming. Smaller parameter may not generate enough force to remove the fragmented material on the surface. For the \texttt{penetrate} primitive, a large parameter could also cause jamming and a smaller parameter will fail to penetrate in the vertical direction. The \texttt{sweep} primitive also consists of sweeping distance limit and contact force limit. If the limit is reached, the sweep will change direction.

\subsection{Offline Reinforcement Learning for Excavation Penetration}

\begin{figure}
	\centering
    \includegraphics[scale=0.32]{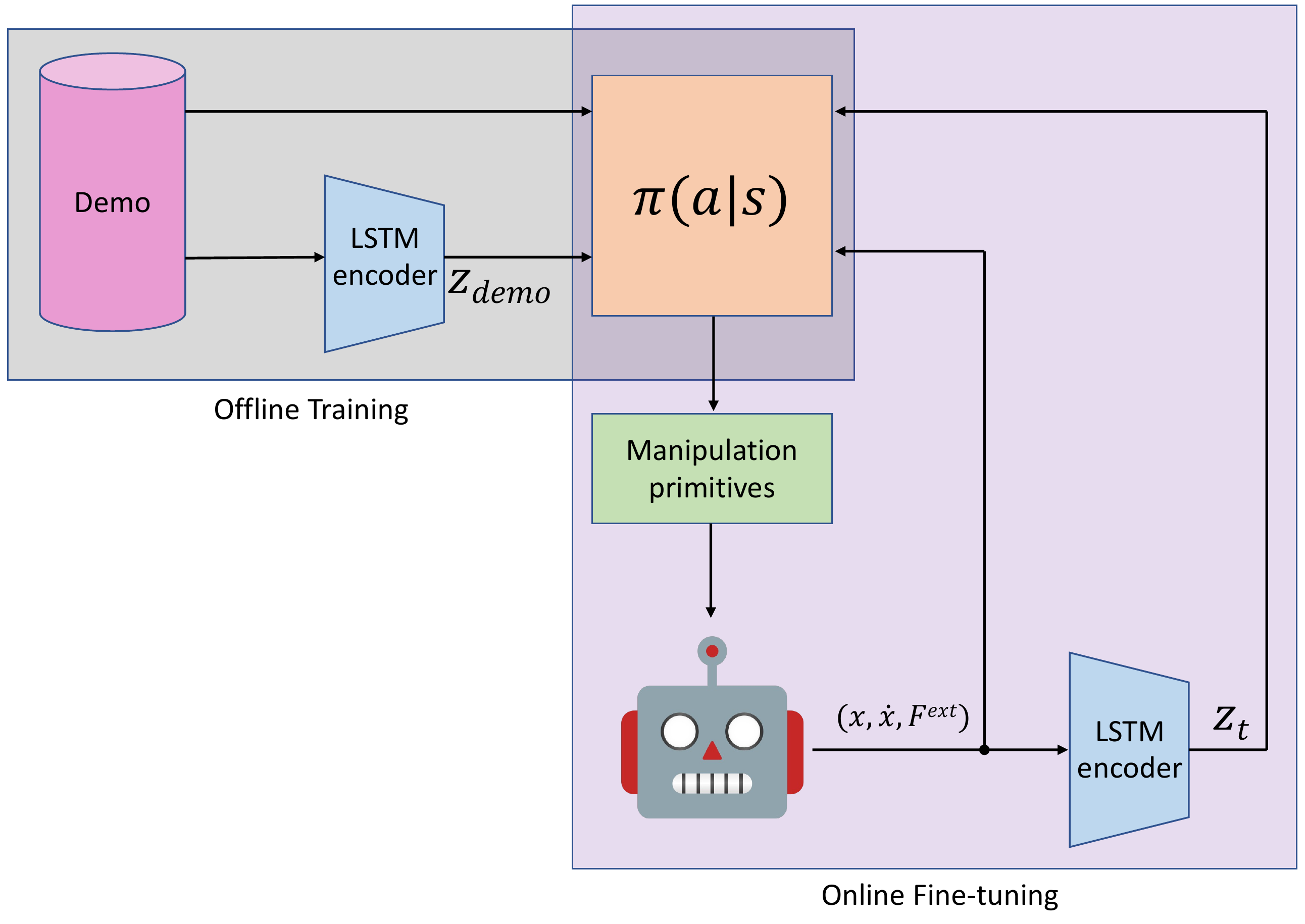}
    \caption{The proposed framework to learn excavation policy using offline RL followed by online fine-tuning.} 
    \label{fig:framework}
\end{figure}

In order to avoid the dangerous and lengthy on-robot learning, we propose to learn the parameters of the manipulation primitives with offline RL from an offline dataset (Fig. \ref{fig:framework}). The dataset consists of demonstrated penetration trajectories collected using teleoperation and scripted policy. For the teleoperation, we manually control the robot bucket using a remote controller. For the scripted policy, we design a rule-based penetration strategy to select the velocities and limits in the primitives. The demonstrated penetration trajectories are saved for offline RL training.  

An optimal penetration strategy should depend on not only the bucket pose and contact force at the current time step, but also the terrain configurations and terrain types. Thus our proposed method consists of two LSTM auto-encoders to extract such features from penetration trajectories. The first encoder network $q_{\phi_1}(z_t|c_{1:t})$ predicts $z_t$ from the current penetration trajectory $c_{1:t} \in \mathbb{R}^{9 \times t}$, where $c_{i} = (x_i, \dot{x}_i, F_i^{ext}) \in \mathbb{R}^{9}$ includes the relative bucket pose $x_i = (x_i^x, x_i^z, \omega_{i}^{pitch}) \in \mathbb{R}^3$ between the measured bucket pose and the initial bucket pose, bucket velocities $\dot{x}_i = (\dot{x}_i^x, \dot{x}_i^z, \dot{\omega}_{i}^{pitch}) \in \mathbb{R}^3$, and measured contact forces $F_i^{ext} = (F_i^x, F_i^z, M_{i}^{pitch}) \in \mathbb{R}^3$ at time step $i$. The second encoder network $q_{\phi_2}(z_{demo}|c_{demo})$ predicts $z_{demo}$ from the demonstrated penetration trajectories $c_{demo} \in \mathbb{R}^{9 \times t_{demo}}$, where $t_{demo}$ is the length of the demonstrated trajectory. Intuitively, with the LSTM auto-encoders, $z_t$ extracts the penetration progress and the current terrain configuration, and $z_{demo}$ extracts the types of rigid objects and the information necessary to excavate these rigid objects. 

The observation space for the penetration policy $\pi (a|s)$ is defined as $s = (x, \dot{x}, F^{ext}, z_t, z_{demo})$.
%where $x$ is the relative bucket pose between the measured bucket pose and the initial bucket pose, $\dot{x}$ is the bucket velocity, and $F_{ext}$ is the measured contact force. 
The action space is defined as the continuous parameters in the manipulation primitives $a = (v^x, F_{lim}^x, d_{lim}^x, \omega^{pitch}, M_{lim}^{pitch}, \alpha_{lim}^{pitch}, v^z, F_{lim}^z)$. The reward function is defined as
\begin{equation}
    r = - w_1 \times ||d_{target}- d||_2^2 - w_2 \times ||F^{ext}||_2^2 
\end{equation}
% $r = - w_1 \times ||d_{target}- d||_2^2 - w_2 \times ||F_{ext}||_2^2 $
where $w_1$ and $w_2$ are the weighted scalars, and $d$ is the measured penetration depth. Essentially, this reward function encourages the bucket to penetrate fast and with small contact forces.

% We learn the penetration policy using offline RL on a user-provided dataset. The dataset consists of penetration trajectories from random actions, tele-operated penetration, and scripted policy penetration. The scr.. tele.. rand... The state includes current bucket pose, bucket velocity, contact force, and a latent variable, which uses LSTM to encode the past penetration trajectory. As the bucket sweep the surface of the terrain during penetration, the latent variable extracts the shape of the terrain and the information necessary to determine how to perform the task for the current terrain shape. 

To learn penetration policy from some sub-optimal demonstrations, we use IQL \cite{kostrikov2021offline}, which has been shown to perform well on tasks that benefit from multi-step dynamic programming by stitching several sub-optimal trajectories. IQL tries to estimate the maximum Q-value over actions that are in support of the data distribution. By utilizing the Expectile Regression, IQL does not need to query the learned Q-function on out-of-sample actions. For our excavation penetration task, the demonstration dataset contains several sub-optimal penetration trajectories. We utilize IQL to stitch those trajectories and find the best trajectory that can finish the task. In another word, the optimal penetration trajectory is composed of several sub-trajectories from the demonstrations. More details about IQL can be found in \cite{kostrikov2021offline}.

\subsection{Online Fine-tuning and Different Terrains Generalization}

Assume there are $N$ training terrains and $M$ unseen terrains. In our case, $N = 5$ and $M = 1$ (fragmented rocks). We can obtain $N$ offline dataset $(\mathcal{D}_1,\mathcal{D}_2,\mathcal{D}_j,...,\mathcal{D}_N )$, where $\mathcal{D}_j$ denotes the $jth$ dataset for the $jth$ terrain. The penetration policy of the $jth$ terrain $\pi_j(a|s)$ can be obtained by training IQL using $\mathcal{D}_j$. After training, $\pi_j(a|s)$ may already find better penetration trajectories than the trajectories in $\mathcal{D}_j$. In addition, $\pi_j(a|s)$ can be improved with a small amount of online interaction. We continue to train IQL by initializing the policy with $\pi_j(a|s)$. The newly collected online interactions are appended to $\mathcal{D}_j$.

Since we include $z_{demo}$ into the observation to extract the types of rigid objects, we can train a general penetration policy $\bar{\pi}(a|s)$ using the entire dataset $ \mathcal{D}_{all} = (\mathcal{D}_1,\mathcal{D}_2,\mathcal{D}_j,...,\mathcal{D}_N )$ \cite{zhao2022offline}. The latent variable $z_{demo}$ performs as a classifier to distinguish different terrains and it also captures the common features that can be generalized to similar terrains. Given an unseen terrain, $z_{demo}$ can be first inferred using $q_{\phi_2}(z_{demo}|c_{demo})$ by only collecting a few demonstrated trajectories. We can then online fine-tune the policy with $\bar{\pi}(a|s)$ as the initialization. This obviates the need for collecting the entire dataset for a new terrain.

\section{Experiments}

We aim to investigate three questions in our real-world experiments. First, we examine if the proposed framework using offline RL can learn the efficient bucket penetration policy without causing jamming. Second, we evaluate whether the learned penetration policy outperforms the sub-optimal demonstrations in the dataset. Third, we inspect whether one learned penetration policy can be applied to different terrain types and handle unseen terrain.

\subsection{Experimental Setup}
\begin{figure}[H]
    \centering
    \includegraphics[scale=0.46]{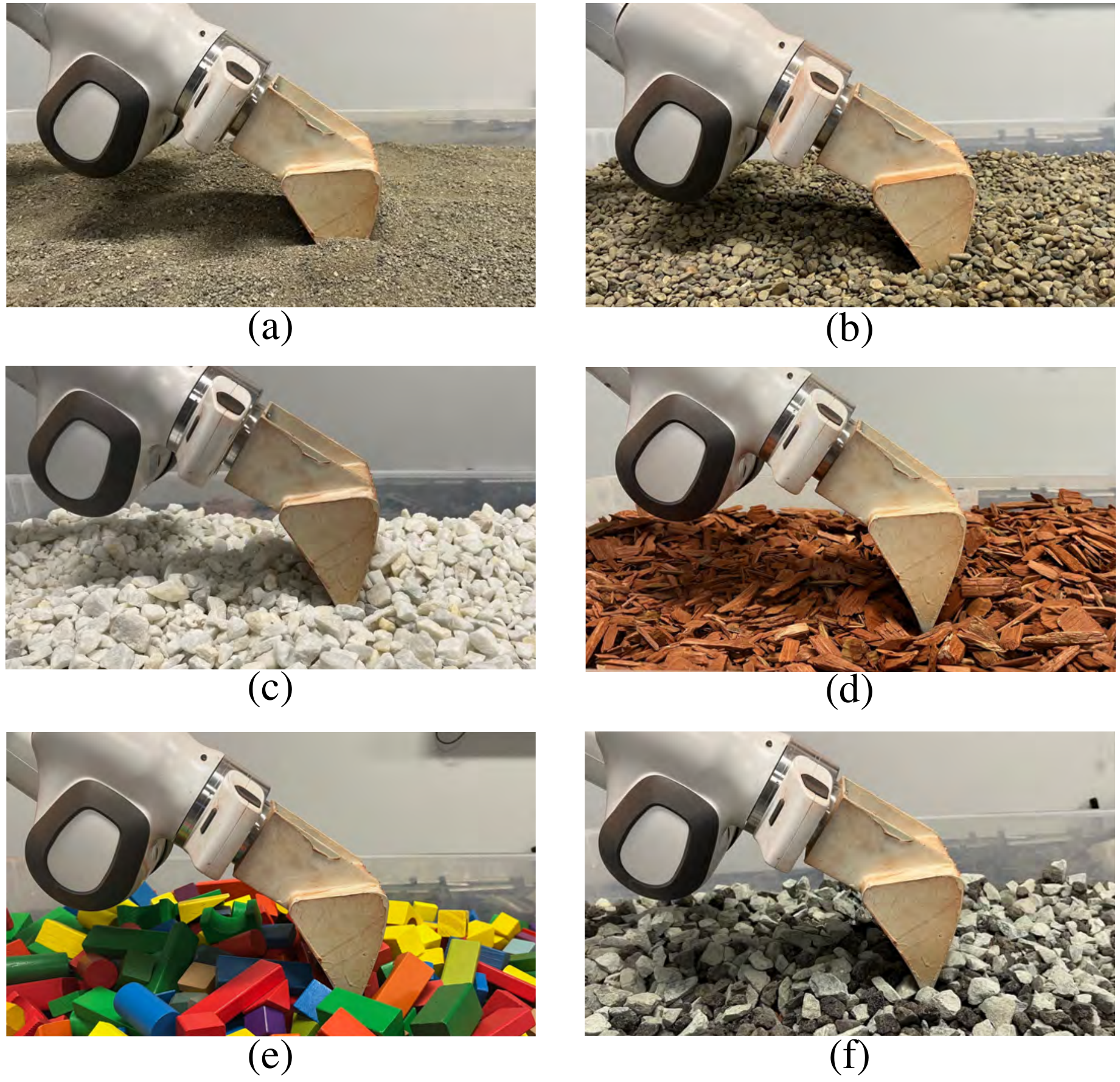}
    \caption{Excavation penetration of 6 different terrains. (a) Sand; (b) Pea Pebbles; (c) Marble Chips; (d) Red Mulch; (e) Wood Blocks; (f) Fragmented Rocks.} 
    \label{fig:rigid_objects}
\end{figure}

As shown in Fig. \ref{fig:setup}, our system includes one 7-DoF Franka robot manipulator, one customized excavator bucket as the end-effector, and one tray containing irregular rigid objects to be excavated. We assume that the objects being excavated are uniformly distributed in the tray with a flat surface. We conducted experiments on 5 different training terrains (sand, pea pebbles, marble chips, red mulch, wood blocks), and one unseen terrain (fragmented rocks), as shown in Fig. \ref{fig:rigid_objects}. To emulate the 4-DoF excavator, we only use the shoulder panning, the shoulder lifting, the elbow lifting, and the wrist lifting joints. The other 3 joints are fixed. The measured bucket-terrain contact force is computed from the joint torques using the Jacobian matrix. The contact force should be measured at the bucket teeth. In the experiments, we use the force at the end-point of the $7th$ link to approximate it. Before penetration, the bucket slowly moves downward to make contact with the terrain. When the contact force is greater than $3N$, we assume that the penetration phase starts. In the penetration phase, the agent controls the bucket at $10Hz$, which is tracked and interpolated by a downstream impedance controller at $1000 Hz$. After a successful penetration, predefined dragging, closing, and lifting trajectories can be executed without causing jamming. Planning the entire excavation trajectory is left to future work.

% And the objects are within the reachable region of the bucket. We assume we can always find a feasible trajectory to excavate the objects. The goal is to penetrate the bucket a certain depth $d$ below the terrain surface without causing jamming. We use $d=5cm$ throughout the experiments. Because we empirically find that $d=5$ is deep enough for full bucket excavation. We also assume that after penetration, the dragging, closing, and lifting trajectories are the same. Planning the entire excavation trajectories are left to future work. In the penetration phase, the agent controls the bucket at 10Hz, which is tracked and interpolated by a downstream impedance controller at 100?? Hz. The bucket contract force is computed from joint torque using Jacobian matrix. 

\subsection{Implementation Details}
\subsubsection{Demonstration Dataset Collection}
We collect offline datasets on 5 training terrains except for the fragmented rocks. For each of the 5 training terrains, 100 trajectories are collected with the scripted policy, and 20 trajectories are collected using teleoperation. In each demonstrated trajectory, there are about 75 $(s, a, r, s')$ transitions. For the scripted policy, we uniformly sample the parameters of the manipulation primitives from the parameter ranges (Table \ref{table1}) as action $a$ . Those ranges are empirically selected to ensure safety for the 5 training terrains and they are also utilized as the action space range in offline RL training throughout the experiments. For the teleoperation, the parameters of the manipulation primitives are extracted from the bucket velocity and the states where the moving direction changes in a demonstrated trajectory. We normalize both the observations and actions space to facilitate the training.
% The fragmented rocks are used to test the generalization ability of the proposed method. 

To add variations in the training dataset, we also uniformly sample the initial bucket-terrain contact position for each trajectory from a $10cm*15cm$ rectangular region on the surface of the terrain. After collecting each trajectory, we manually reset the terrain to a flat surface. Note that the distributions of the irregular rigid objects are different after each reset, thus the contact dynamics are also different.

\renewcommand{\arraystretch}{1.5} % change row height
\begin{table}
\centering
\caption{Parameters range of the manipulation primitives.}
\label{table1}
\begin{tabular}{|c|c|c|}
\hline 
 Primitives & Parameters &  Range \\\hline
 \multirow{3}{*}{\texttt{sweep}}&$v^x$ & $[-0.1, 0.1]m/s$ \\
 &$F_{lim}^x$&$[25, 40]N$ \\
 &$d_{lim}^x$& $[0.015, 0.03] m $\\\hline
 \multirow{3}{*}{\texttt{rotate}}&$\omega^{pitch}$& $[-0.5, 0.5]rad/s $\\
 &$M_{lim}^{pitch}$ & $[40, 70]N \cdot m$\\
 &$\alpha_{lim}^{pitch}$ & $[0.1, 0.2] rad$ \\\hline
  \multirow{2}{*}{\texttt{penetrate}}&$v^z$ & $[-0.03, 0.03]m/s$\\
  &$F_{lim}^z$ & $[40, 70]N$\\\hline
\end{tabular}
\end{table}
% We only collected data for the penetration phase with \texttt{sweep} and \texttt{penetrate} primitives. The data was collected by random actions, teleoperation, and scripted policy. For random actions, random parameters within the safety range are used for the primitives. For teleoperation, we remote control the bucket for penetration. For the scripted policy, we manually select the primitive parameters and force/position limits. For each terrain, we collect ?? trajectories for random actions, ?? trajectories for teleoperation, and ?? trajectories for scripted policy. In each trajectory, there are about 100 $(s, a, r, s')$ transitions. 

\subsubsection{Behavioral Cloning Baseline}
We implement behavioral cloning to learn a baseline policy from the offline dataset to evaluate the learned offline RL policy. The network contains a Multilayer Perceptron (MLP) with two fully connected layers. We use ReLU as the activation function. And the learning rate is $3 \times 10^{-4}$. The inputs and outputs are the observations and actions in the offline dataset, respectively. We use the same auto-encoders in the offline RL framework to encode the trajectories.

\subsubsection{Offline Reinforcement Learning and Online Fine-tuning}
 We use the same training parameters as in the original IQL paper. We use Adam optimizer with a learning rate $3 \times 10^{-4}$, 2 layer MLP with ReLU activations, and 256 hidden units for all networks. 
%The individual policy $\pi_i(a|s)$ is trained using dataset $\mathcal{D}_i$. The general excavation policy $\pi(a|s)$ is trained using all the dataset $(\mathcal{D}_1,\mathcal{D}_2,\mathcal{D}_i,...,\mathcal{D}_N )$.
For the reward function, we empirically find that $d_{target}=0.05m$ is deep enough for the downstream excavation. The weighted scalars $w_1$ and $w_2$ are 400 and 0.0004, respectively.  

For the online fine-tuning, $z_{demo}$ is predicted using 10 trajectories from the offline dataset. If the terrain has not been seen before, such as fragmented rocks, we recollect 10 penetration trajectories using the scripted policy to predict $z_{demo}$. (The reason that we recollect the trajectories using the scripted policy instead of the learned policy is to avoid distribution shifts. More details can be found in \cite{pong2022offline}). In the online fine-tuning phase, we collect 20 trajectories, which is about $1500$ new online interactions.

% We first train two LSTM-VAEs for the trajectories. The first encoder network encodes the entire demonstration trajectory. The latent variables perform as a classifier to distinguish different terrains and it also captures the common features that can be generalized to similar terrains. The second encoder network encodes the current trajectory until current time step. The latent variables extract the current terrain shape and the information necessary to complete the current task. 

% Once we have the two encoder networks, we train the offline RL policy on the collected dataset. For the reward function, we use $r = d - d_{target} + w*||F||_2^2$, where $d$ and $d_{target}$ are the measured and target penetration depth, respectively, $w$ is a weight scalar, and $F$ is the measured contact force. For each terrain, we train the penetration policy using the dataset collected on that terrain. This give us $N$ penetration policies, where $N$ is the number of the terrain types. We also train a general penetration policy using the dataset collected on all the terrains. We evaluate the learned offline policy on each of the terrains.

% \subsubsection{Online fine-tuning}
% We collect 2000 new online interactions
% We also conduct experiments on online fine-tuning the learned RL policy. If we have the offline dataset for the terrain, we will initialize the policy with the learned policy. If the task is unseen, we will initialize the policy with the learned general penetration policy. Online interactions are collected to fine-tuning the policy using IQL.

\subsection{Results and Discussions}

\renewcommand{\arraystretch}{2.0} % change row height
\begin{table*}
\centering
\caption{Real-world experiments results of different penetration policies $\pi_{j}(a|s)$ learned on their own datasets $\mathcal{D}_{j}$.}
\label{table2}
\begin{tabular}{|c|c|c|c|c|c|c|}
\hline 
 Terrains &  \thead{Sand }&  \thead{Pea Pebbles }&  \thead{Marble Chips } & \thead{Red Mulch} & \thead{Wood Blocks} & \thead{Fragmented Rocks} \\\hline
 Vertically Downward &  Jamming  &  Jamming & Jamming & Jamming & Jamming & Jamming \\\hline
 Behavioral Cloning &$-48.90 \pm 11.06$&  $-35.40 \pm 6.72$  &  $-45.23 \pm 10.08$ & $-30.20 \pm 9.96$ & $-25.52\pm 7.82$  & $ - $\\\hline
 Offline RL $\pi_{j}(a|s)$  &  $-39.23 \pm 9.30$  &   $-30.63 \pm 5.49$& $-35.39 \pm 15.12$  &  $-26.84 \pm 6.54$  & $-22.20 \pm 7.94$ 
 & $-$\\\hline
 Fine-tuned $\pi_{j}(a|s)$ &  $-26.76 \pm 4.60$  & $-28.83 \pm 2.50$  & $-30.58\pm 3.30$ &$-27.03 \pm 3.96$  &  $-24.53 \pm 7.09$ & $-$\\\hline
 
\end{tabular}
\end{table*}

\renewcommand{\arraystretch}{2.0} % change row height
\begin{table*}
\centering
\caption{Real-world experiments results of one general policy $\bar{\pi}(a|s)$ learned on the entire dataset $\mathcal{D}_{all}$. }
\label{table3}
\begin{tabular}{|c|c|c|c|c|c|c|}
\hline 
 Terrains &  \thead{Sand }&  \thead{Pea Pebbles }&  \thead{Marble Chips } & \thead{Red Mulch} & \thead{Wood Blocks} & \thead{Fragmented Rocks} \\\hline
 General Policy $\bar{\pi}(a|s)$ &  $-28.51 \pm 7.02$  & $-32.37 \pm 7.72$ & $-36.14\pm9.46$  &  $-57.90 \pm 25.94$  &   $-25.37 \pm 8.32$
 & $-40.17 \pm 11.17$\\\hline
 Fine-tuned $\bar{\pi}(a|s)$ & $-26.07 \pm 3.48$   &  $-29.50 \pm 4.01$ & $-31.11\pm7.29$ & $-23.02 \pm 5.81$  &  $-23.98 \pm 6.60$ & $-37.05 \pm 10.81$\\\hline
 
\end{tabular}
\end{table*}

%%%%%%%%%%%%%%%%%%%%%%%%%%%%%%%%%%%%%%%%
%%%%%%%%%%%%%%%%%%%%%%%%%%%%%%%%%%%%%%%%

% \renewcommand{\arraystretch}{2.0} % change row height
% \begin{table*}
% \centering
% \caption{Real-world experiments results of different penetration policies on different terrains.}
% \label{table2}
% \begin{tabular}{|c|c|c|c|c|c|c|}
% \hline 
%  Terrains &  \thead{Fixed \\trajectory }&  \thead{Behavioral \\cloning }&  \thead{Offline RL \\ $\pi_{i}(a|s)$ } & \thead{Fine-tuned \\$\pi_{i}(a|s)$} & \thead{General policy\\$\pi(a|s)$} & \thead{Fine-tuned\\$\pi(a|s)$} \\\hline
%  Sand &  no jamming  &  $-48.90 \pm 11.06$ & $-39.23 \pm 9.30$ & $-26.76 \pm 4.60$ & $-28.51 \pm 7.02$ & $\mathbf{-26.07 \pm 3.48}$ \\\hline
%  Pea pebbles &jamming&  $-35.40 \pm 6.72$  &  $-30.63 \pm 5.49$ & $\mathbf{-28.83 \pm 2.50}$  & $-32.37 \pm 7.72$& $-29.50 \pm 4.01$ \\\hline
%  Marble chips &  jamming  &  $-45.23 \pm 10.08$ & $-35.39 \pm 15.12$ & $\mathbf{-30.58\pm 3.30}$ & $-36.14\pm9.46$ & $-31.11\pm7.29$\\\hline
%  Red mulch &  jamming  &  $-30.20 \pm 9.96$ & $-26.84 \pm 6.54$ & $-27.03 \pm 3.96$ & $-57.90 \pm 25.94$& $\mathbf{-23.02 \pm 5.81}$ \\\hline
%  Wood blocks &  jamming  &  $-25.52\pm 7.82$ & $\mathbf{-22.20 \pm 7.94}$ & $-24.53 \pm 7.09$ & $-25.37 \pm 8.32$ & $-23.98 \pm 6.60$\\\hline
%  Fragmented rocks &  jamming & 
%  %$-60.37\pm13.83$ 
%  N/A& N/A & N/A & $-40.17 \pm 11.17$ & $\mathbf{-37.05 \pm 10.81}$\\\hline

% \end{tabular}
% \end{table*}
\subsubsection{Offline Reinforcement Learning Evaluation}

Table \ref{table2} shows the performance of different penetration policies $\pi_{j}(a|s)$ on different terrains. Note that the rewards are negative. A larger reward means better penetration performance. There are 5 training terrains with offline RL policies. To evaluate the performance of the policies, we execute the policy on the terrain for 10 trials in each scenario. We then record the means and standard deviations of the trajectory reward in the table. No results are shown for the fragmented rocks using behavioral cloning, offline RL $\pi_{j}(a|s)$, and fine-tuned $\pi_{j}(a|s)$, because no offline datasets are collected on the fragmented rocks. 

We also show the result of vertically downward trajectories, where the bucket follows a fixed trajectory to move vertically downward. 5 out of 6 terrains have jamming problems due to the large contact force. In contrast, the learned offline RL policies do not cause jamming. We also compare the performance of behavioral cloning and offline RL $\pi_{j}(a|s)$ in Table \ref{table2}. Offline RL outperforms behavioral cloning on all 5 terrains. This shows the effectiveness of the learn offline RL policies.
% This suggests that offline RL can learn a better penetration policy from the pre-collected dataset, even if the dataset only contains sub-optimal trajectories. 
By fine-tuning RL $\pi_{j}(a|s)$, we can see that 3 out of 5 learned offline RL penetration policies are improved with a few online interactions.
% , with one exception for the wood blocks. For the wood blocks, all 5 learned policies perform similarly. The reason is that the wood blocks have relatively large sizes and light weight. The robot does not need to apply large forces to move those blocks during the penetration. 

\begin{figure}[H]
    \centering
    \includegraphics[scale=0.3]{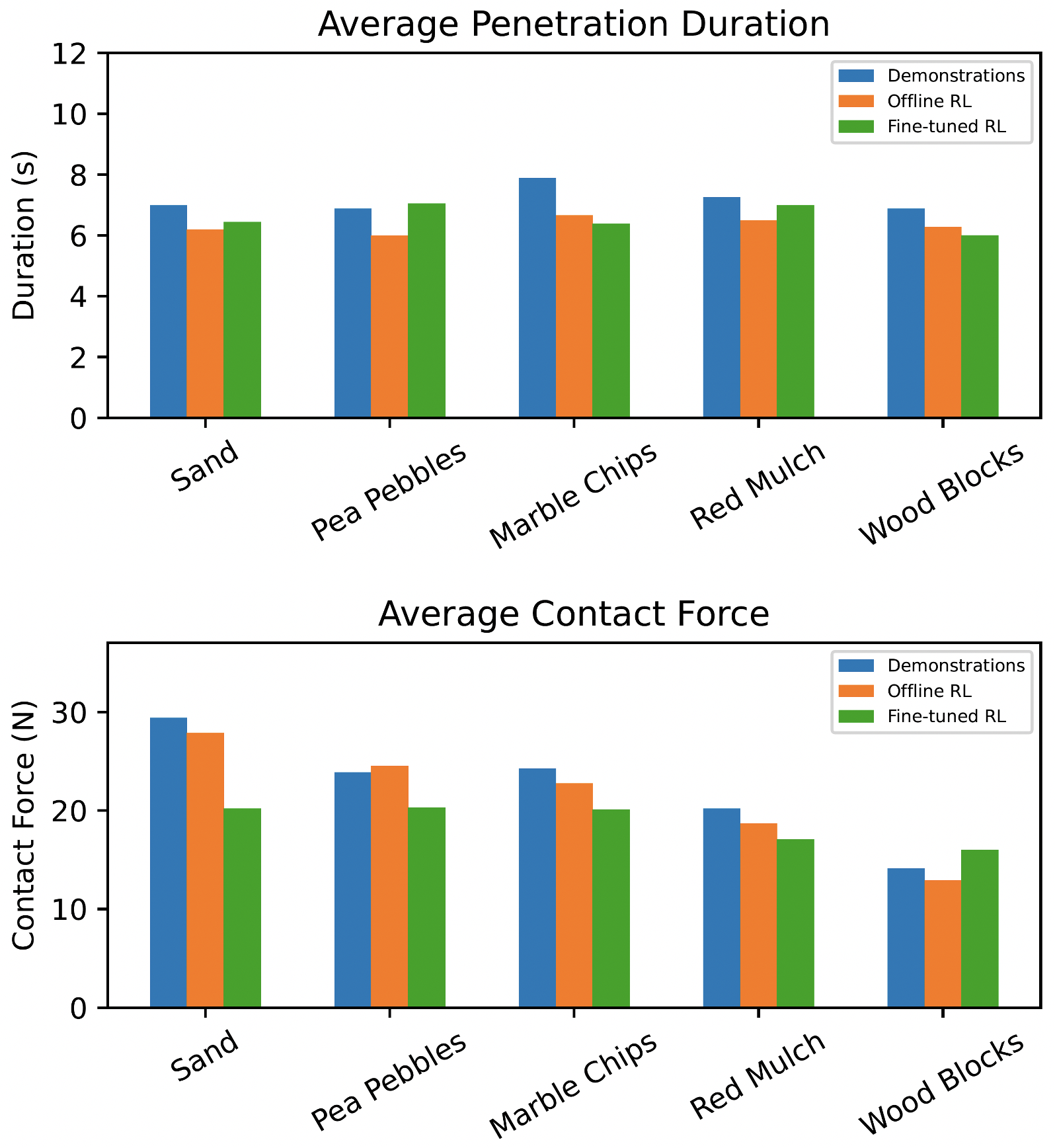}
    \caption{Average penetration duration and contact force for the demonstrations and the RL policies.} 
    \label{fig:average_force_duration}
\end{figure}

Since the numbers in reward may not be intuitive to humans, we also report the average penetration duration and contact force in a trajectory (Fig. \ref{fig:average_force_duration}). As we can see that the learned offline RL policies have faster penetration and smaller contact forces than the demonstration trajectories. This suggests that the RL policies outperform the demonstrations in the dataset. There are some exceptions, for example, the average contact force in fine-tuned RL for wood blocks is larger than the average contact force in the demonstrations. The possible reason is that the contact force for wood blocks penetration is relatively small due to the light weight of wood blocks. So the contact force would take smaller weights in the reward function, and the RL policies sacrifice contact force for faster penetration.

To evaluate the generalization ability of the proposed method, we show the performance of one general policy $\bar{\pi}(a|s)$ trained on the entire dataset $\mathcal{D}_{all}$ (Table \ref{table3}). 
% Note that Table \ref{table2} also reports the performance of the learned general policy $\bar{\pi}(a|s)$. 
% This single policy $\bar{\pi}(a|s)$ achieves better performance than behavioral cloning. 
This suggests that one learned penetration policy can be successfully applied to different rigid objects. The last row in Table \ref{table3} shows that the general policy can also be improved with online fine-tuning. There is one outlier data point for the red mulch. The possible reason is that the red mulch is long and thin. One particle of red mulch got stuck into the bucket teeth in the experiments, resulting in a large contact force during the penetration. 
% The last column in Table \ref{table2} shows that the general policy can also be improved with online fine-tuning.
% and reach a similar performance to fine-tuned $\pi_{j}(a|s)$. 
For the unseen fragmented rocks, we consider it the hardest task. Because the particles of the fragmented rocks are larger and sharper than the pea pebbles and the marble chips. As shown in the last column in table \ref{table3}, the learned $\bar{\pi}(a|s)$ performs well on the fragmented rocks without causing jamming, suggesting that $\bar{\pi}(a|s)$ is able to handle out-of-distribution rigid objects. 

\subsubsection{A Case Study for Changing the Jamming Threshold }
\begin{figure}
    \centering
    \includegraphics[scale=0.69]{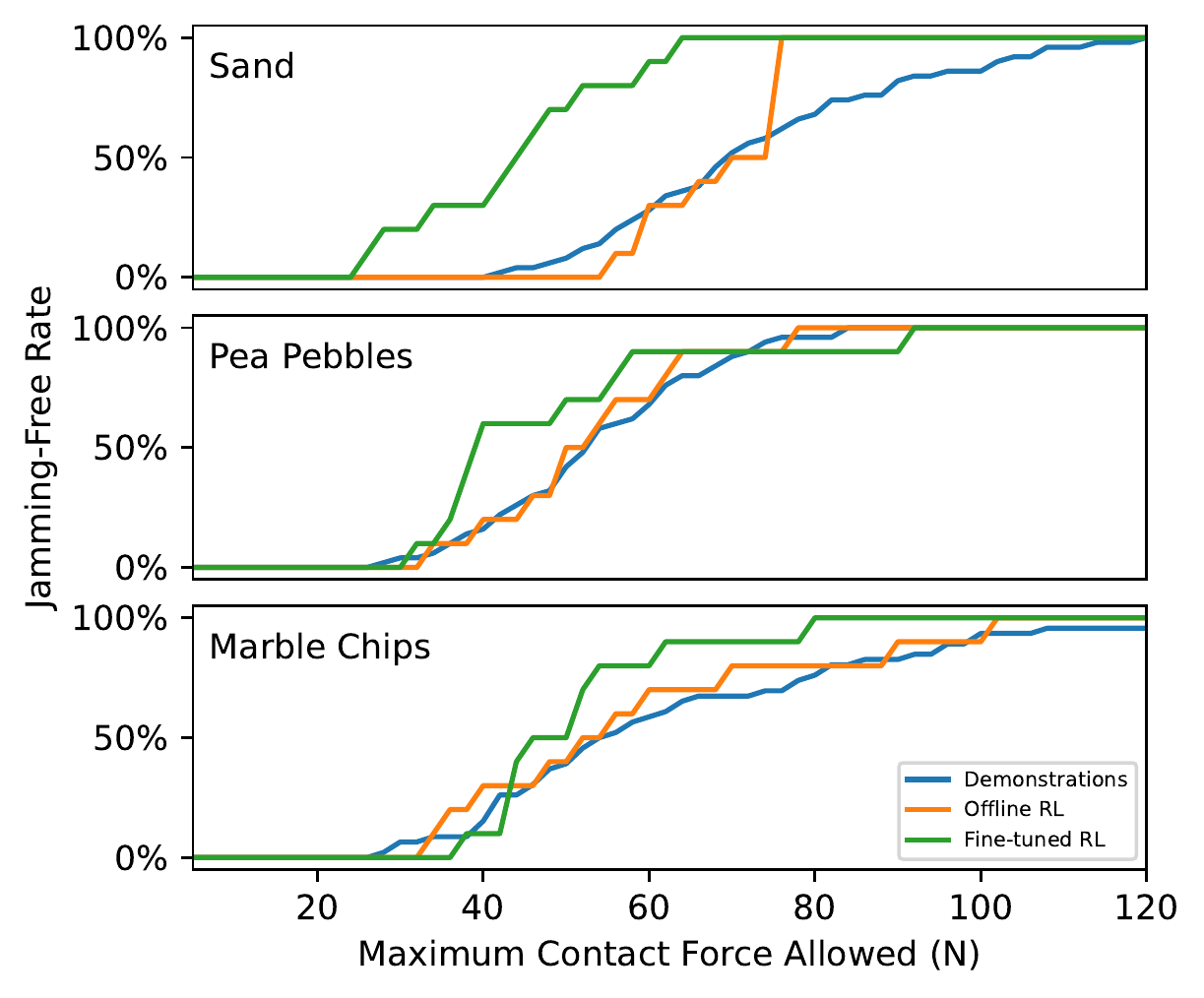}
    \caption{Jamming-Free Rate VS Maximum Contact Force Allowed} 
    \label{fig:jammint_free_rate}
\end{figure}

As mentioned in Section \ref{problem_formulation_section}, the Franka robot controller will issue a halt to the robot if large contact forces are detected. This is utilized as a criterion for whether jamming happens or not. However, different robots or excavators could generate different maximum torques. A jamming-free scenario for one robot may become a jamming problem if we only have the access to another less powerful machine. If we redefine the jamming threshold as the maximum contact force allowed in one penetration trajectory, we can plot the jamming-free rate versus the maximum contact force allowed in Fig. \ref{fig:jammint_free_rate} (We have recorded the maximum contact force in a trajectory when we evaluate each trajectory). In the figure, as the maximum contact force allowed increases, the jamming-free rate also increases. We can see that the learned offline RL policies have higher jamming-free rates than the demonstrations. This shows the advantages of the learned offline RL policies when the jamming criteria become more strict.

\section{Conclusion and Future Work}
This work proposes an offline reinforcement learning framework for the robotic excavation of rigid objects. We learn the excavator bucket penetration policies from the offline collected dataset. Three manipulation primitives are proposed and the action parameters are learned with offline RL.
% The proposed framework can successfully deal with the jamming problem in bucket penetration. 
Real-world experiments show that the learned policy can successfully avoid jamming problems and outperform the sub-optimal demonstrations in the dataset. One learned policy can also be applied to different terrains and quickly adapt to an unseen terrain with only a few online interactions. One of the limitations of this work is the lack of visual perception. It will be helpful to utilize the camera to find the gaps between rigid objects before penetration starts, especially for large rocks and stones. For future work, it is noteworthy that our framework only focuses on the penetration phase in the excavation process. More general excavation policies are possible to be learned using this framework if we provide a corresponding offline demonstration dataset for other excavation phases. 
% We would also like to incorporate visual perception to increase the robustness of the excavation policy. Compared to the bucket size, the rigid objects that we are handling in this paper are relatively small. The excavation of large rigid objects is still an open problem. Finally, we would like to evaluate our proposed method on a full-size excavator. 

%                                   % the textheight of the last page by a suitable amount.
%                                   % This command does not take effect until the next page
%                                   % so it should come on the page before the last. Make
%                                   % sure that you do not shorten the textheight too much.

\bibliographystyle{IEEEtran}
\bibliography{reference}

\end{document}